\documentclass{article}

\usepackage{microtype}
\usepackage{graphicx}
\usepackage{booktabs} 

\usepackage{times}
\usepackage[T1]{fontenc}
\usepackage{natbib}
\usepackage{bm,amsmath,amssymb,amsthm}

\usepackage{hyperref}
\hypersetup{colorlinks=true,urlcolor=blue,citecolor=red}
\providecommand{\url}[1]{}

\providecommand{\Comment}{\COMMENT}
\providecommand{\EndFor}{\ENDFOR}
\providecommand{\For}{\FOR}
\providecommand{\RETURN}{\STATE \textbf{return} }
\providecommand{\State}{\STATE}
\providecommand{\Statex}{\ \newline }
\providecommand{\EndProcedure}{\ }
\providecommand{\Call}[2]{\textsc{#1}(#2)}
\providecommand{\Procedure}[2]{\textsc{#1}(#2)}

\providecommand{\True}{\texttt{true}}

\DeclareMathOperator{\Real}{\mathbb{R}}

\DeclareMathOperator{\tr}{\mathrm{tr}}

\DeclareMathOperator{\sign}{\mathrm{sign}}
\DeclareMathOperator{\expect}{\mathbb{E}}
\DeclareMathOperator{\variance}{\mathbb{V}}

\newcommand{\code}{\texttt}
\newcommand{\send}{\code{send} }

\newcommand{\normal}[2]{\mathcal{N}\left(#1,#2\right)}

\newtheorem{problem}{Problem}

\renewcommand{\vec}[1]{{\boldsymbol{#1}}}
\newcommand{\x}{\vec{x}}

\newcommand{\toencrypt}{\stackrel{\text{encrypt}}{\longmapsto}}
\newcommand{\todecrypt}{\stackrel{\text{decrypt}}{\longmapsto}}

\newcommand{\define}{\stackrel{\cdot}{=}}

\newtheorem{lemma}{Lemma}
\newtheorem{definition}{Definition}

\newcommand{\enc}[1]{{\left\langle \! \left\langle #1\right \rangle \! \right\rangle}}
\newcommand{\dec}[1]{{\left\langle \! \left\langle #1\right \rangle \! \right\rangle^{-1}}}

\usepackage[accepted]{icml2018}

\icmltitlerunning{Private Text Classification}

\begin{document}

\twocolumn[

\icmltitle{Private Text Classification: Using Radomacher operators and blind addition to allow multiple peers to learn text without seeing the text}

\icmlsetsymbol{equal}{*}

\begin{icmlauthorlist}
\icmlauthor{Leif W. Hanlen}{}
\icmlauthor{Richard Nock}{}
\icmlauthor{Hanna J. Suominen}{}
\icmlauthor{Neil Bacon}{}
\end{icmlauthorlist}


\icmlcorrespondingauthor{Leif Hanlen}{leif.hanlen@gmail.com}

\icmlkeywords{Machine Learning, Private, Text Analytics, NLP}

\vskip 0.3in
]

\begin{abstract}
Confidential text corpora exist in many forms, but  do not allow arbitrary sharing. 
We construct typical  text processing applications using appropriate privacy preservation techniques (including homomorphic encryption, Rademacher operators and secure computation). We set out the preliminary materials from Rademacher operators for binary classifiers, and then construct basic text processing approaches to match those binary classifiers.
 %


\end{abstract}

%
%
%
%

\footnotetext{Leif W. Hanlen is with CSIRO, Data61, and Australian National University and University of Canberra.
Richard Nock and Neil Bacon are with CSIRO, Data61. 
Hanna J. Suominen is with CSIRO, Data61,  University of Canberra,  Australian National University, University of Turku, and University of Canberra.
}

\section{Motivation}

%


Private text data --- with confidential  content --- is difficult to ``open''. Privacy requirements in text data are difficult to guarantee due to the inter-dependencies of text, and grammar. Although \textit{Natural Language Processing} (NLP) nominally operates on numerically encoded text, NLP exploits  the structure of text and not merely a sequence of integer codes~\cite{Hirschberg:2015aa}.
 
Research work in the space of \textit{Information Retrieval }(IR) has acknowledged the need to preserve privacy~\cite{Si:2014zi,Oard:2015aa}.  
Mechanisms to accommodate sharing of text corpora, have essentially reduced to licensing requirements that allow full access under limited conditions of (re-)use, since sharing (raw) text data essentially allows a human to  read (and reproduce) the  data~\cite{Khaled-El-Emam:2013ye,Thomson:2004ij,Ji:2014hb,JISC:2012hz}.

%
%
\subsection{Machine Learning from Private Text Data }

The (typical) components of the text document that might be  subject to privacy concerns (names of persons, places, drug-names, disease-names) are likely to be the  components most interesting for text processing  --- and most damaging to algorithm performance if altered. 
In this work, we consider a different approach: we apply  encryption techniques to text which  allow learning without viewing the raw data --- thereby applying machine learning without the need to share (or read) text data to
   %
learn from private text corpora and 
classify private text. 


%
\emph{This work does not avoid the need for ethical approval and research permission: encryption and privacy preserving techniques cannot overcome ethical, legislative, or contractual requirements on what may  (or may not) be done with  data. We are interested in allowing groups to ethically interact   with   data, where raw data sharing would not be desired (or possible). }
%

\subsubsection{Text in Health Is a Special Case}


 %
 %
Although open sharing is generally accepted as a  good principal in health~\cite{Verhulst:2014rr,Dunn:2012,Estrin:2010,Veitch:2010};  privacy concerns may overwhelm  implied scientific benefit~\cite{Thomson:2004ij,Vogel:2011,Sayogo:2012fr}. 
 %
 %
The need to address privacy while supporting research-use (as well as non-research use) of health data has been observed~\cite{McKeon:2013kx}.
To support confidentiality, the British Medical Journal~[Table 1~\cite{Hrynaszkiewicz:2010qr}  recommends \emph{not} publishing verbatim responses or transcriptions of clinical discussions --- which is exactly the sort of data that text mining systems  require~\cite{Suominen:2014px,Suominen:2014rw}.
The work of~\cite{Jin:2007aa} suggests some approaches for privacy-preserving health analytics, and reviews several privacy-preserving techniques, although most are numeric~focused.


The risk of compromising privacy by being able to conceal the identifiers remains regardless of recent advances in automated de-identification algorithms for health text.
Algorithms for automated de-identification of health text have 
have been evaluated to reach the F1 correctness percentage
from 81 to 99 in English, French, Japanese, and Swedish~\cite{DalianisVelupillai2010,Moritaetal2013,Chazardetal2013,GrouinNeveol2014,Kayaalpetal2014,Meystreetal2014}.\footnote{F1 is a performance measure that takes values between 0 and 1 --- the larger the value, the better the performance. It is defined as the harmonic mean of precision and recall, that is, $2 \times (\textrm{precision} \times\textrm{recall}) / (\textrm{precision} + \textrm{recall})$ where  precision refers to the proportion of correctly identified words for de-identification to all de-identified words and recall refers to the proportion of correctly identified words for de-identification to all words that should have been de-identified.} However, approximately
90 per cent of the residual identifiers left behind by either these algorithms or human coders can be
concealed by applying additional computation methods~\cite{Carrelletal2013}. 

This capability to conceal the identifiers gets even more alarming 
after record linkage of different shared corpora. For example, in the USA, Washington is one of 33 states that share or sell anonymized patient records. For US\$50, anyone can purchase a patient-level health corpus that contains all hospitalisations that occurred in this state in 2011, without patient names or addresses, but with full patient demographics, diagnoses, procedures, attending physician, hospital, a summary of charges, and how the bill was paid. 
By linking these de-identified health records with public news papers from the same year from Washington State, 
leads 43 per cent  of the time to concealing the patient's name and sometimes even her/his street address~\cite{Sweeney2015},

As expected, patients are
concerned about the potential of health data sharing and linkage to result in data misuse and compromised
privacy~\cite{Simonetal2009}. 
However, they are also enthusiastic about their
capacity to improve the quality and safety of health care through 
giving their informed consent to sharing some or all of their 
own health records for purposes of (medical) science in general or some specific research project~\cite{Shawetal2015}.

Our approach addresses precisely these problems in finding and ``hiding'' sensitive text. It allow machine learning algorithms to use all encrypted data, but not all raw text.





\section{Background}

 %
We assume  all participants are \textit{Honest But Curious }(HBC)~\cite{Paverd:2014aa}.
 %
We limit the need for ``trusted'' intermediaries~\cite{Dwork:2014aa}, and where such intermediaries are used, we restrict (by aggregation and secure computing) the information they may receive. 

\subsection{Text Processing}\label{SS:text:pipeline}

We  consider the \emph{typical} linear  so-called ``1-best'' pipeline as outlined in~\cite{Johnson:2014ab}. This could be extended to parallel, iterative, or network approaches. 
 %
 %
 %
We shall ignore feature engineering and simply presume a (large) number of categorical variables. 
A particular example of text processing pipeline output is outlined in Figure~1 of~\cite{Hirschberg:2015aa}. 
 %
%
%

In training, the text processing pipeline also uses labelled text segments --- which may be document-, sentence- or word- labels (e.g. for sentiment analysis) or some combination of text fragments (such as used by the \textit{Browser Rapid Annotation Tool }(BRAT) for collaborative text annotation~\cite{Stenetorp:2012aa}). In each case, we may  represent the features as numeric labels --- where the ``tags'' are converted into a dictionary --- and a series of numeric values. We shall be interested in binary values --- such as might result from a ``1-hot'' encoding.
These become our  observations, and also labels.


 A differential-privacy approach is not suitable for this data type: adding ``noise'' to the  encoded text will either render the ``new'' text meaningless, or be  overcome --- by treating the noise as   spelling or grammatic errors. 
 %
 %
We  use the  approach of~\cite{Shannon:secrecy} --- to improve  independent secret systems by concatenating them. In this case, we will firstly encrypt  the numeric features and labels (using a Paillier homomorphic encryption system~\cite{Paillier:1999aa,Damgoard:2001qc}). This makes direct interpretation difficult. Second, we attempt to address the data dependencies by applying irreversible aggregation to the numeric data so as to hide many of the implied dependencies between observations. Finally, we wrap the learning process in a secure learning approach, to further reduce the capacity of an inquisitive user discovering the underlying labels. 
This reflects the well known fact that data dependences must  be   accounted for in training, validation, and testing of machine learning methods in order to produce reliable performance estimates~\cite{Suominen:2008aa,Pahikkala:2012aa}.


\begin{figure*}
\centering
\includegraphics[width=\textwidth]{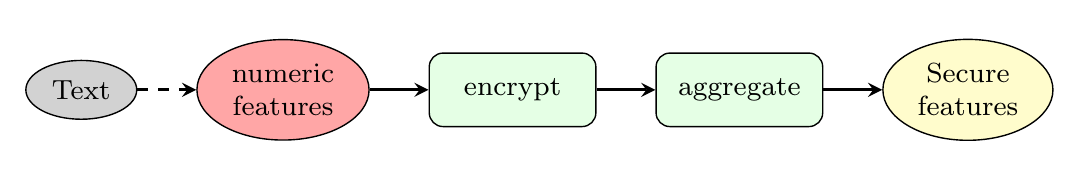}
\caption{Text processing pipeline with encryption,   pre-processing   is encapsulated in the dashed arrow.}\label{F:pipeline:encrypt}
\end{figure*}


\subsection{Partial Homomorphic Encryption} \label{SS:homomorph}

The Paillier encyption scheme~\cite{Paillier:1999aa} (and later generalisations~\cite{Damgoard:2001qc}) is a public-private key encryption scheme. We alter the notation of~\cite{Franz:2011aa} (note, this is different to~\cite{Djatmiko:2014bs}) where an integer $x$ is encrypted as $\enc{x}$ and the decrypt operation is $\dec{y}$. In other words, 
\begin{equation}\label{endecrypt}
x \stackrel{\text{encrypt}}{\longmapsto} \enc{x} \stackrel{\text{decrypt}}{\longmapsto} \dec{\enc{x}} = x.
\end{equation}
 
The operations $\enc{\cdot}$ and $\dec{\cdot}$ in Definition (\ref{endecrypt}) are public key encryptions: users can encrypt data (and perform computations) using a common public key, however, only the user with the corresponding private key can extract the clear data.

The main operations for Paillier homomorphic encryption are the operators $\oplus$ and $\otimes$. They are defined for two integers $x_1,x_2<n=pq$ where $n$ is a constant for the particular encryption and  parameters $p$ and $q$ are two large primes as follows:
\begin{equation}\label{E:paillier.sum}
\enc{x_1} \oplus \enc{x_2} = \enc{x_1+x_2}
\end{equation}
and
\begin{equation}\label{E:paillier.multiply}
\alpha \otimes \enc{x_2} = \enc{\alpha \cdot x_2}
\end{equation}
where $\alpha$ is an \emph{un-encrypted} real-valued scalar.

Equations \eqref{E:paillier.sum}  and \eqref{E:paillier.multiply} gain us an ability to 
sum encrypted values \emph{in the encrypted domain} (and consequently decrypt the result) and multiply encrypted values with un-encrypted scalars. Note that the result \eqref{E:paillier.multiply} does not apply when  $\alpha$ is  encrypted. 
For more advanced operations (such as multiplying encrypted values) we use secure computation.

\subsection{Secure Computations}\label{SS:secure}

We use results from~\cite{Franz:2011aa,From:2006aa} to provide several protocols for secure computation among two parties $\mathcal{A}$ and $\mathcal{B}$. Work from \cite{Clifton:2002aa} provides mechanism for multiple parties (i.e., more than two).
 %
We shall assume that $\mathcal{A}$ operates on encrypted data, and $\mathcal{B}$ has the private key (and can decrypt data). Neither party should be able to discern the numerical values. 
 %
These protocols comprise the following three  steps:
\begin{enumerate}
\item an obfuscation step by  the public key holder $\mathcal{A}$, 
\item a  transfer step to the private key holder $\mathcal{B}$,  who decrypts and then performs the calculation on clear data\footnote{As the result is obfuscated, $\mathcal{B}$ learns nothing from this operation, even though it is performed on clear data.} and returns  an encrypted result to $\mathcal{A}$, and
\item $\mathcal{A}$ then removes the original obfuscation.
\end{enumerate}
 %

Other work extends~\cite{Franz:2011aa} to linear algebra for homomorphic analysis.
 %
 %


We now recall the work of \cite{Nock:2015aa} and \cite{Patrini:2015ab} to present relevant parts the aggregation techniques.
This presents learners on specially aggregated data sets where the data set \emph{could} be in a single location. 

\subsubsection{Single (Complete) Data Set}

We will first consider the data set as a single (coherent) source.
That is, all data is held by a single~organisation. 

\begin{definition}[(Numeric) Supervised Learning  Space]
Given a set of $m>0$ examples  $\mathcal{S} = \left\{(\vec{x}_i,y_i), i\in \{1,2,\ldots,m\} \right\}$, where $\vec{x}_i\in\mathcal{X}\subseteq\Real^{1\times d}$ are observations, $\mathcal{X}$  is the domain, and $y_i \in \{-1,1\}$ are binary labels. We are concerned with a (binary) linear classifier $\vec{\theta}\in \Theta$ for fixed $\Theta\subseteq\Real^{1\times d}$. The label of an observation $\vec{x}$ is given by
\begin{equation*}
\mathrm{label}(\vec{x}) =\sign\left( \vec{\theta}^T \vec{x}\right)\in\{-1,1\}.
\end{equation*} 
\end{definition}

The irreversible aggregation is based on \textit{Rademacher observations} (rados) as defined below:

\begin{definition}[Rado] cf.Definition 1~\cite{Nock:2015aa}

Let $\Sigma=\{-1,1\}^n$. Then given a set of $\mathcal{S}$, and for any $\vec{\sigma}\in\Sigma$ with $\vec{\sigma} = [\sigma_1, \ldots, \sigma_n]^T$. The Rademacher observation $\vec{\pi}_\vec{\sigma}$ with signature $\vec{\sigma}$ is
\begin{equation}\label{E:rado}
\vec{\pi}_\vec{\sigma} = \frac{1}{2} \sum_{i=1}^n (\sigma_i +y_i) \vec{x}_i.
\end{equation}
\end{definition}


%
%

\subsubsection{Multiple Data Sets}\label{SSS:multiple}

%
%
%

This case is described in Figure~1~\cite{Patrini:2015ab}. We do not assume that entities are linked: different text corpora are held by different parties, and no entity resolution is performed. 


\begin{definition}[BB-Rado] cf. Definition~1~\cite{Patrini:2015ab} 
Consider $\vec{z'}\in U\subset V$. Let $\mathrm{lift}(\vec{z'})$ concatenate zeros  to $\vec{z} = [\vec{z'} \ \vec{0}]$ such that $\vec{z}\in V$. For any $\vec{s}\in\mathcal{J}$, labels $y\in\{-1,1\}$ and $\alpha\in\Real$, the $\alpha$-basic block rado for $(\vec{s},y)$ is
\begin{multline}\label{E:bb-rado}
\vec{\pi}^{\alpha}_{(\vec{s},y)} \define \alpha \cdot \mathrm{lift}(y\cdot \vec{s}) + \sum_{j=1}^p \mathrm{lift}\left(\vec{\pi}^j_{(\vec{s},y)}\right).
\end{multline}
\end{definition}

\subsection{Encrypted Rados} \label{SS:secure:rado}


The encryption process occurs \emph{after} securely processing the text documents at the private location of $\mathcal{B}$. Using her/his private key, $\mathcal{B}$ then encrypts the features, and these are then aggregated. The aggregation occurs blind to $\mathcal{B}$, and may be performed by an honest-but-curious intermediary $\mathcal{I}_\mathcal{B}$. The rados $\vec{\pi}_\vec{\sigma}$ are generated privately at $\mathcal{I}_\mathcal{B}$. Once generated, the rados can be used by other honest-but-curious external parties $\mathcal{A}$. 

\begin{figure*}
\centering
\includegraphics[width=\textwidth]{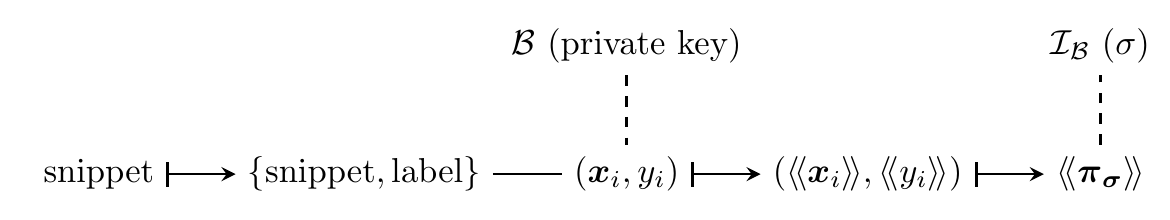}
\caption{Feature encryption pipeline, showing encryption links and (dashed) knowledge at $\mathcal{B}$ with private key, and intermediary $\mathcal{I}_{\mathcal{B}}$.}\label{F:encrypt}
\end{figure*}

Figure~\ref{F:encrypt} outlines the encryption steps, 
using secure mathematical operations, and denote the two parties as $\{\mathcal{B},\mathcal{I}\}$ where $\mathcal{B}$ is the private key holder and $\mathcal{I}$ is an intermediary. 
 %
$\mathcal{I}$ can ``see'' \emph{encrypted} features $\enc{\vec{\x}_i}$,  and encrypted labels $\enc{y_i}$, and knows the choices of rado vectors (i.e., $\mathcal{I}$ knows values of $\sigma_i$). It would be possible to operate with $\sigma_i$ \emph{also} encrypted. 

%
%

We re-write Equation \eqref{E:rado} below with the encrypted values made explicit. Corresponding secure mathematical operations are also shown. We use the notation $\bigoplus_{i=1}^n$ to denote a series of homomorphic addition operations ie. $\bigoplus_{i=1}^n = a_1 \oplus a_2 \oplus \cdots \oplus a_n$. We will use $:$ as an abuse of notation, to denote ``has the meaning of'' rather than equality, as follows:

\begin{equation}\label{E:rado:meaning}
\enc{\vec{\pi}_\vec{\sigma}} \quad : \quad \frac{1}{2} \sum_{i=1}^m \left(\sigma_i +\enc{y_i}\right) \enc{\vec{x}_i}.
\end{equation}

The resulting ``Equation''~\eqref{E:rado:meaning} shows the formation of the  (encrypted) rado. The additions and (unencrypted scalar) multiplications must all be translated to the appropriate homomorphic addition and  multiplication operations. 

The output is an encoded rado, based on any numerical field, that we will refer to as $\enc{\vec{\pi_\sigma}}$. We outline the procedure to build the rado in Protocol~\ref{A:homo-rado}: \nameref{A:homo-rado}.

\begin{algorithm}
\floatname{algorithm}{Protocol}
\caption{
Encrypted Radomacher}\label{A:homo-rado}
\begin{algorithmic}
\ENSURE $\enc{\vec{\pi_\sigma}}$
\State at peer
\State $\vec{\sigma} \toencrypt \enc{\vec{\sigma}}$
\State $\enc{\vec{\phi}} \gets \enc{\vec{\sigma}} \oplus \enc{\vec{y}}$
\State $\enc{\vec{\phi}} \gets\frac{1}{2} \otimes \enc{\vec{\phi}}$
\State  $\enc{\vec{\pi_\sigma}} \gets $ \Call{S.InnerProd}{$\enc{\vec{\phi}}, \enc{\vec{x}}$} 
\end{algorithmic}
\end{algorithm}

\subsubsection{Multi-party Case}

The case for multiple parties requires the use of the $\mathrm{lift}(\cdot)$ function.  This function appends zeros onto a vector, and thus (in the encrypted domain) may be represented as appending an encrypted scalar (zero) to the encrypted vector. As above, Equation \eqref{E:bb-rado} can be re-written in the encrypted~domain.

\subsection{Learning, Using Encrypted Rados}\label{SS:secure:learn}


\subsubsection{Unencrypted Single-party Case}
Recall the learner for rados (in the unencrypted case) is given by \cite{Nock:2015aa}. 
We will use the equivalent (exponential)  learner for rados as follows:

\begin{lemma}[Rado Learning](cf.~Lemma~2~\cite{Nock:2015aa} )
For any $\vec{\theta}$ and $\mathcal{S}$, and a set of  rados \mbox{$\vec{\sigma}\in\mathcal{U}\subseteq\Sigma$}, minimizing the loss 
\begin{equation}
\log\left[F_{\exp}(\mathcal{S},\vec{\theta})\right] = \log \left[ \frac{1}{n}\sum_{\vec{\sigma}\in\mathcal{U}} \exp\left(-\vec{\theta}^T \vec{\pi}_\vec{\sigma} \right)\right]
\end{equation}
is equivalent to  minimising the standard logistic loss $F_{\log}(\mathcal{S}, \vec{\theta})$.
\end{lemma}

The supervised learning (optimisation) is written as
\begin{problem}[Minimise Exponential Loss]\label{P:exploss}
The optimal classifier $\vec{\theta}^*$ is given by solving
\begin{equation}
\vec{\theta}^* = \min_\vec{\theta}J(\vec{\theta})
\end{equation}
where
\begin{equation}\label{E:J}
J(\vec{\theta})= \log \left[ \frac{1}{n}\sum_{\vec{\sigma}\in\mathcal{U}} \exp\left(-\vec{\theta}^T \vec{\pi}_\vec{\sigma} \right)\right] + \vec{\theta}^T\vec{\theta}
\end{equation}
and $ \vec{\theta}^T\vec{\theta}$ is a regularising term (a.k.a. regulariser).
\end{problem}


\subsubsection{Secure Single-party Case}


The exponential in Equation \eqref{E:J} can be computed securely using the protocol outlined in~\cite{Yu:2011aa}. The logarithm can be performed using Algorithm~1~\cite{Djatmiko:2016aa}. 
 %
 %
We perform a gradient descent to solve Problem~\ref{P:exploss}.


Recall Problem~\ref{P:exploss}. 
Note that the gradient of $J(\vec{\theta})$, with respect to $\theta_j$, is
\begin{equation}
\frac{\partial}{\partial \theta_j} J(\vec{\theta}) =
 \frac{ \sum_\vec{\pi} -\pi_j \exp\left(-\sum_i \theta_i \pi_i \right) }  %
 { \sum_\vec{\pi} \exp\left(-\sum_i \theta_i \pi_i \right) } 
+
2\theta_j.
\end{equation}

We note that $\enc{\vec{\pi}}\in \mathcal{P}$. 

Using our abuse of notation $:$ we have
\begin{equation}
\enc{J( {\vec{\theta}}) } \ : \ \left[ \sum_j -\pi_j \exp\left(-\sum_i \theta_i \pi_i \right) \right].
\end{equation}

 %
%



\subsubsection{Unencrypted, Multi-party Case}

The proof of Theorem~3~\cite{Patrini:2015ab} shows that the mean square loss can be used, over $\mathcal{P}$ --- that is, on the limited sample sets --- by using a modified mean loss as given in Definition~\ref{D:bb-rado} as follows: 
\begin{definition}[BB-Rado loss]\label{D:bb-rado} cf. Definition~2~.\cite{Patrini:2015ab}  and Theorem~3~\cite{Patrini:2015ab}  
The $M$-loss for the classifier $\vec{\theta}$ is
\begin{multline}\label{E:bb-rado-short}
\ell_M(\mathcal{R}_{S,\mathcal{P}}) \define -\left(\expect_\mathcal{P} \left[ \vec{\theta}^T \vec{\pi}_\vec{\sigma} \right] 
- \frac{1}{2}\variance_\mathcal{P} \left[  \vec{\theta}^T \vec{\pi}_\vec{\sigma}\right] \right)
\\+\frac{m}{4} \vec{\theta}^T \Gamma \vec{\theta}
\end{multline}
where expectation $\expect_\mathcal{P}$ and variance $\variance_\mathcal{P}$ are computed with respect to the uniform sampling of $\sigma$ in $\mathcal{P}$. If the matrix $\Gamma$ is positive definite, it can be defined as a weighted diagonal matrix
\begin{equation}
\Gamma = \begin{bmatrix}
 I_m & \vec{0}_m \\
\vec{0}_n &\epsilon I_n
\end{bmatrix}
\end{equation}
where $0 < \epsilon \ll 1$ accounts for (lack of) confidence in certain columns of $\vec{\pi}$.
\end{definition}

\emph{``[M]inimizing the Ridge regularized square loss over examples is equivalent to minimizing a regularized version of the M-loss, over the complete set of all rados.''}~\cite{Patrini:2015ab} 

The optimal classifier $\vec{\theta}^\star$ is given by the simple closed-form expression Theorem~6~\cite{Patrini:2015ab}. Namely,
\begin{equation}\label{E:theta-opt-patrini}
\vec{\theta}^\star = \left(BB^T + \mathrm{dim}_c(B)\cdot \Gamma\right)^{-1} B \vec{1}
\end{equation}
where $B$ is stacked (column-wise) rados and $\mathrm{dim}_c(B)$ is the number of columns of $B$.
The procedure for building the rados and solving for $\vec{\theta}$ are given in \cite{Patrini:2015ab}.



To solve  \eqref{E:theta-opt-patrini}, \cite{Hall:2013aa} recommends  an iterative -- Shur~\cite{Guo:2006aa}) approach.
 %
Faster approaches (with fewer multiplications) are achieved by higher order algorithms. An outline and review are given in~\cite{Rajagopalan:1996aa,Soleymani:2012aa}.
The inverse may be found using secure multiplication linear algebra.

%


\subsubsection{De-risking the Coordinator}

The notation of~\cite{Patrini:2015ab} suggests a central coordinator with access to all vectors,: we  avoid this  by returning to Definition~\ref{D:bb-rado}.
 %
%
Let 
\begin{equation}\label{E:sum:b}
\vec{b} = \sum_{\vec{\pi}\in\mathcal{P}} \vec{\pi} = \expect_\mathcal{P}\left(\vec{\pi}_\vec{\sigma}\right)
\end{equation}
and then
\begin{equation}\label{E:sum:def:theta}
\frac{\partial \ell}{\partial \vec{\theta}} =
-\vec{b}^T +  \vec{\theta}^T\left[ \Gamma + \frac{1}{2}\sum_{\vec{\pi}\in\mathcal{P}} \left(\vec{\pi} -\vec{b} \right) \left(\vec{\pi} -\vec{b} \right)^T\right].
\end{equation}


The sums in Equations \eqref{E:sum:b} and \eqref{E:sum:def:theta} are over the appropriate rados. However, these rados may be calculated by their peers, so the sums 
may be broken into per-peer summations, where we consider disjoint sets $\mathcal{P}_p$ such that $\mathcal{P}_1\cup \mathcal{P}_2\cup \ldots =\mathcal{P}$.

\begin{definition}[BB-rado per peer]
Consider $P$ peers with $p\in\{1,2,\ldots,P\}$, where each peer $p$ has distinct  rados drawn from $\mathcal{P}_p$, and the rados are distinct in $\mathcal{P}_1\cup \mathcal{P}_2\cup \ldots = \mathcal{P}$.  For each peer $p$, we have a expectation   $\vec{e}_p$ and variance  $\vec{v}_p$ defined as
\begin{equation}
\vec{e}_p = \sum_{ \mathcal{P}_p} \vec{\pi}_i 
\end{equation}
and
\begin{equation}
\vec{v}_p = \sum_{\mathcal{P}_p} \left(\vec{\pi}_i -\vec{b} \right) \left(\vec{\pi}_i -\vec{b} \right)^T. 
\end{equation}
Each peer $p$ can calculate $\vec{e}_p$ and  $\vec{v}_p$  independently.

\end{definition}

Although the mean $\vec{b}$ \emph{may} be calculated centrally, it is preferable to use secure multi-party addition to achieve the same result. This reduces the scope for the coordinator to access (encrypted, aggregated) vectors, and (instead) only access noisy aggregates of the data.

\section{Putting the Bricks together}



The algorithm incorporates the secure multi-party summation work of~\cite{Clifton:2002aa}, to prevent the coordinator from obtaining the rados directly. This  adds a third layer of obfuscation to the data (encryption, aggregation, blind addition), which means that at the coordinator (who can decrypt the data) the data remains protected by the aggregation to rados and the blind addition of the~vectors.

\begin{algorithm}
\floatname{algorithm}{Protocol}
\caption{
Classifier for Secure Text with Central Coordinator}\label{A:centralcontol}
\begin{algorithmic}
\REQUIRE peers $p\in\{1,\ldots,P\} $, 
  coordinator $\mathcal{C}$

\ENSURE encrypted classifier $\enc{\vec{\theta}}$ at  $\mathcal{C}$
\ENSURE encrypted local classifier  $\enc{\vec{\theta_p}} $ at peer $p$ 
\ENSURE binary feature vector $\vec{f_p}$  at  $\mathcal{C}$ \Comment{ the features  available at each peer $p$}


\Statex at  coordinator $\mathcal{C}$:
\State generate Paillier public key  $\mathbf{K}$  \& secret key $\mathbf{S}$  as a pair
\State \send public key $\mathbf{K}$ to all peers $p\in\{1,\ldots,P\}$

\Statex at each peer $p$ independently: 
\Statex \Comment{run local text  labelling on document set} 
\Statex \Comment{$D_p$ document set;
$\mathbb{D}$ dictionary; 
$\{X_p,\vec{y_p}\}$ data and labels } 
\State $\{X_p,\vec{y_p}\} \gets $ \Call{LocalText}{$\mathbb{D}, D_p$}
\State $\vec{f_p} \gets$ binary vector of observations
\State \send $\vec{f_p}$ to $\mathcal{C}$
\State encrypt data and labels $\{X_p,\vec{y_p}\} \toencrypt  \left\{\enc{X_p} , \enc{\vec{y_p}}\right\}$
\State  build rado's  from encrypted data $\enc{\vec{\pi}_i}$ using public key $\mathbf{K}$
\State  $\enc{\vec{\pi}_{(p)}} \gets \bigoplus_i \enc{\vec{\pi}_i}^{\mathbf{K}}$
\Comment{elementwise homomorphic addition}

\Statex at  coordinator $\mathcal{C}$:
\State $\enc{\vec{b}}\gets$ \Call{SM.Add}{$\{1,\ldots,P\}, \left\{ \enc{\vec{\pi}_{(p)}}  \right\}$} 
 \Comment{\textsc{Sec.Add} using $\enc{\vec{\pi}_{(p)}}$}
\State \send mean value $\enc{\vec{b}}$ to all peers $p=\{1,\ldots,P\}$

\Statex at each peer $p$ independently:
\State  $\enc{\vec{u}_i} \gets \enc{\vec{\pi}_i} \oplus (-1)\otimes \enc{\vec{b}}$ 
\State  $\enc{B_i} \gets \Call{S.OuterProd}{p,\mathcal{C}, \enc{\vec{u}_i},  \enc{\vec{u}_i}^T }$ \Comment{\textsc{Sec.outerProduct}}
\State  $\enc{B_{(p)}} \gets \bigoplus_i B_i$ \Comment{elementwise homomorphic addition}
\Statex at  coordinator $\mathcal{C}$:
\State $\enc{A}\gets$ \Call{SM.Add}{$\{1,\ldots,P\}, \left\{ \enc{B_{(p)}}   \right\}$} 
 \Comment{\textsc{Sec.add} using $\enc{B_{(p)}} $}
\State $\enc{V} \gets$ \Call{S.Inv}{$\enc{A} \oplus \enc{\Gamma}$} \Comment{\textsc{Sec.inversion}}
 \State $\enc{\vec{\theta}} \gets$ \Call{S.MatProd}{$\enc{V}$,$\enc{\vec{b}}$} \Comment{\textsc{Sec.mult}}
\For{$p=1$ to $P$}
\State $\enc{\vec{\theta_p}} \gets \enc{\vec{\theta}\left[\vec{f_p \equiv \True}\right]} $\Comment{local classifier for  peer}

\State \send $\enc{\vec{\theta_p}} $ to peer $p$ 
\EndFor


\end{algorithmic}
\end{algorithm}

\begin{figure*}[t]
\centering
\includegraphics[width=\textwidth]{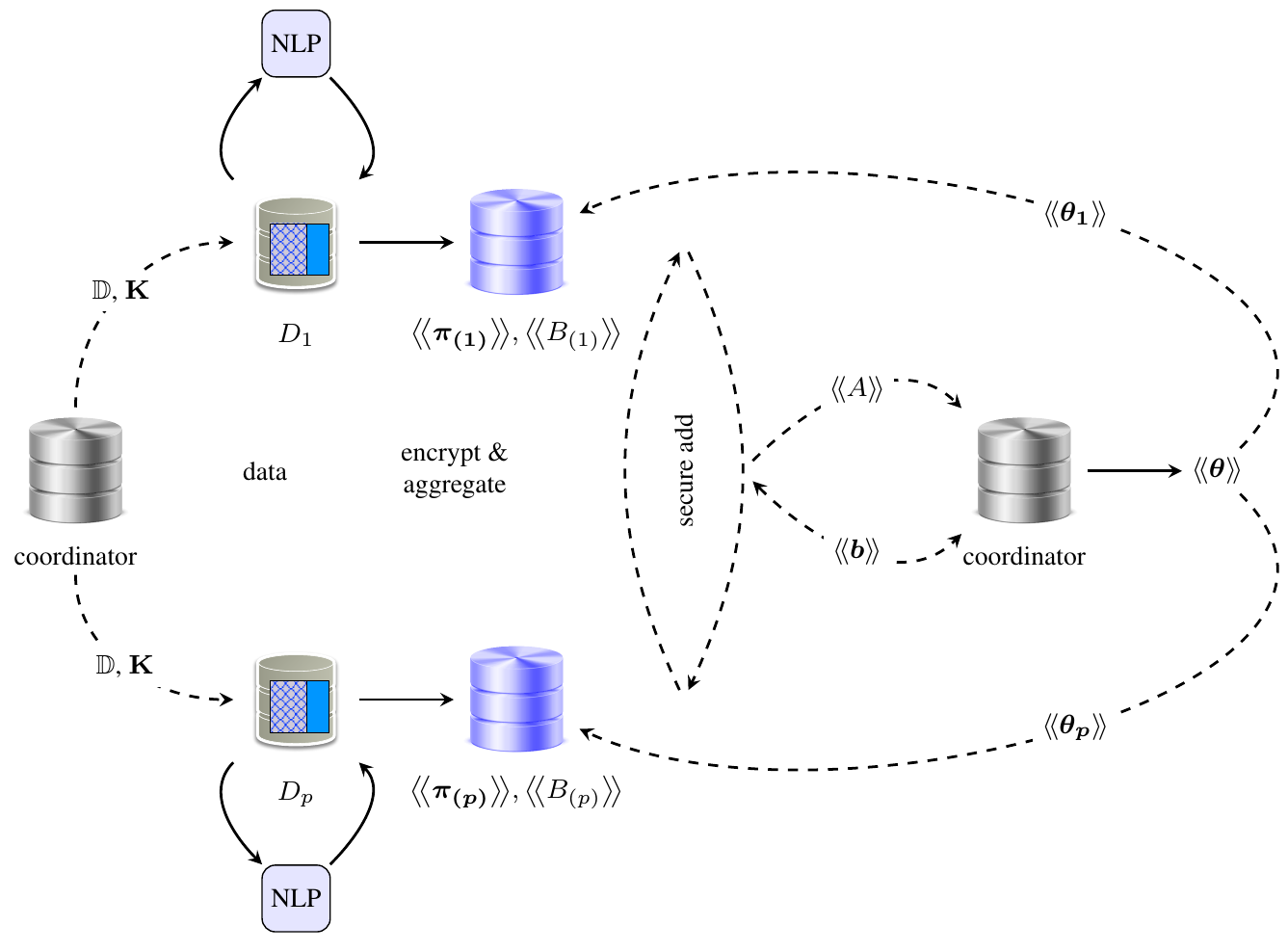}
\caption{Communication architecture for multiple peers, common coordinator. 
The coordinator sends a common dictionary and public key to all peers. Each peer has different data components, with some common elements (cyan).
Each peer has  encrypted and aggregated its local data (blue server icons). The blue servers  correspond to the ``intermediary'' in Figure~\ref{F:encrypt}.
The encryption key is generated by the coordinator. Dashed arrows denote information transfers between participants, whilst solid arrows denote local transformations (at the respective participant). 
 %
}\label{F:server-setup-nock} 
\end{figure*}


In Figure~\ref{F:server-setup-nock} we have outlined the encrypted pipeline, that combines  Figure~\ref{F:pipeline:encrypt} with the inverse proposed in~\cite{Patrini:2015ab}, using Protocol~\ref{A:centralcontol}: \nameref{A:centralcontol}.

At each peer $p$, we now wish to classify a particular observation vector $\vec{x}$. Nominally, we would~calculate 
\begin{equation}
\hat{y} = \sign \left( \vec{\theta}^T \vec{x} \right).
\end{equation}

However, each peer only has a subset of features $\vec{x}_{[:]}$. We note that the label is determined \emph{only} by the sign of a scalar, and hence, we can break the inner product  $\vec{\theta}^T \vec{x}$ into an inner product of local features and remote features as follows:
\begin{align}
\hat{y}_p &= \sign \left( 
\vec{\theta}_{\text{local}}^T  \vec{x}_{\text{local}}
+
\vec{\theta}_{\text{remote}}^T  \vec{x}_{\text{remote}}
\right)
\\
&= \sign \left( 
\underbrace{\vec{\theta_p}^T  \vec{x_p}}_{\text{local}}
+
\underbrace{\sum_{q\neq p}
\vec{\theta_q}^T  \vec{x_q}}_{\text{remote}}
\right). \label{E:local.remote.sign.class}
\end{align}
The local component of Equation \eqref{E:local.remote.sign.class} may be calculated at  peer $p$. If we denote the local classifier result as $\alpha_p$, then we may write
\begin{align}
\alpha_p &= \vec{\theta_p}^T \vec{x_p} \textrm{ and }
\\
\hat{y} &=  \sign \left(  \sum_{p} \alpha_p\right)
\\
 &=\sign \left( \alpha_p + \sum_{q\neq p} \alpha_q\right) \label{E:local.remote.sign.class.2}
\end{align}

The summation in Equation \eqref{E:local.remote.sign.class.2} is the sum of all (local) calculated classifier results on the sub-components of the vector $\vec{x}$.
The result of Equation \eqref{E:local.remote.sign.class.2} shows that the remote  classification results may be treated as \emph{offsets} for the local result --- that is, the remote inner products act as corrections to the local result. However, this requires that every peer share linking information about the observation $\vec{x}$. To avoid this, we replace the  summation in Equation \eqref{E:local.remote.sign.class.2} with an equivalent rado as follows:
\begin{equation}
\hat{y}=\sign \left( \alpha_p + \vec{ \theta_{\neg p}}^T \vec{\pi} \right). \label{E:local.remote.sign.class.2}
\end{equation}

In the homomorphic encrypted case, the local inner  product can be calculated by keeping the encrypted classifier vector $\enc{\vec{\theta}}$ in its encrypted form, and treating  the elements of $\vec{x}$ as unencrypted scalars. 
Finally, the summation may be achieved using multi-party secure addition, as outlined in~\cite{Clifton:2002aa}.


\begin{algorithm}
\floatname{algorithm}{Protocol}
\caption{
Local Classify for Secure Text}\label{A:local.classify}
\begin{algorithmic}
\REQUIRE   coordinator $\mathcal{C}$ with public-secret key pair $\mathbf{K}_C,\mathbf{S}_C$
\REQUIRE   common extra rado $\enc{\vec{\pi}}$ at $\mathcal{C}$
 
\REQUIRE binary feature vector $\vec{f_p}$  at  $\mathcal{C}$  \Comment{ the features  available at each peer $p$}
\REQUIRE encrypted local classifier  $\enc{\vec{\theta_p}} $ at each peer $p\in\{1,\ldots,P\}$ 

\ENSURE $\hat{y}$ label from classification
\Statex at  peer $p$:
\State $\enc{\alpha_p} \gets $ \Call{S.local.innerProd}{$p,\vec{x}, \enc{\vec{\theta_p}}$} \Comment{\textsc{Local.scalarproduct}}
\Statex at each \emph{other} peer $q\neq p$: \Comment{The set of peers may be chosen by $\mathcal{C}$}
\State $\enc{\alpha_q} \gets $ \Call{S.innerProd}{$q,\mathcal{C},\enc{\vec{\pi}}, \enc{\vec{\theta_q}}$} \Comment{\textsc{Sec.innerproduct}}

\Statex at  peer $p$:
\State $\enc{\alpha} \gets $ \Call{SM.Add}{$p=\{1,\ldots,P\}, \{\enc{\alpha_p}\}$}
\Comment{\nameref{A:secure-add} using scalar $\enc{\alpha_p}$}
\State \send $\enc{\alpha}$ to $\mathcal{C}$
\Statex at $\mathcal{C}$:
\State $\enc{\alpha} \todecrypt \alpha$
\State \send $s=\sign(\alpha)$ to peer $p$

\Statex at $p$:
\State $\hat{y} \gets s$
\end{algorithmic}
\end{algorithm}

\begin{table*}[t]
\centering
\caption{Results using numerical regression, and trivial text analytics. Encryption does not impact the accuracy of the results, but does dramatically reduce computation speed.}\label{T:results}
\begin{tabular}{|l|cccrr|}
\hline
Grad descent algorithm &peers &rados  &$\theta$ &misclassification &run time (s)\\
\hline
LogisticRegression (baseline) &1 &plain &plain &0.049 &0.13s\\
radoBoost 
&1 &plain & plain
&0.12
&1.4
\\
radoLearn using rados from radoBoost 
&1 &plain & plain
& 0.20 &0.069
\\
\hline
radoLearn &4  &plain &plain &0.089 & 0.12
\\
radoLearn  &4 &encrypted &plain &0.10 &75 \\
radoLearn &4 &encrypted &encrypted &0.085 &87\\
\hline
\end{tabular}
\end{table*}

\subsection{Usage Scenario for Multi-party, Private Text}
In this scenario we outline the key procedure for learning from distributed private text corpora, and then classifying a locally private corpora. We shall use names to illuminate actors. Alice and Bob  each have private text collections. Alice would like to classify her text by using a combination of patterns learnt from her own data and from Bob's.  Cate\footnote{Cate plays no role in the learning, but is needed as a coordinator.} provides coordination for Alice and Bob. Together, Alice, Bob and Cate follow Protocol~\ref{A:centralcontol} to establish a learned feature set. As Alice and Bob may have different feature sets, Cate separates Alice's appropriate feature vector, and sets the remaining features to zero. Cate then coordinates Alice and Bob through Protocol~\ref{A:local.classify}.

\section{Preliminary results}

Using a simple data set from UCI Ionosphere data set -- to provide a significant number of numeric features, agnostic of text input -- we compare basic analytics using various privacy constraints. For comparison,  
\cite{Zhou:2004ab} has reported a misclassification rate of  0.109, 0.112 \& 0.096  using various neural network approaches. In our case, we have trialled standard linear regression, against multiple peers with rados, and $\theta$ all calculated in the encrypted domain. The results are shown in Table~\ref{T:results}

%

\begin{algorithm}
\floatname{algorithm}{Protocol}
\caption{Secure Rado Solver}\label{A:secure-solve}
\begin{algorithmic}
\STATE $\enc{T} \gets$ \Call{S.MatProd}{$\left\{\enc{B},\enc{B}^T\right\}$} \COMMENT{\textsc{Sec.mult}}
\STATE $\enc{T} \gets \enc{T} \oplus \left[ \mathrm{dim}_c(B) \otimes \enc{\Gamma} \right]$
\STATE $\enc{V} \gets$ \Call{S.Inv}{$\left\{\enc{T}\right\}$} \COMMENT{\textsc{Sec.inversion} }
\STATE $\enc{V} \gets$ \Call{S.MatProd}{$\left\{\enc{V},\enc{B}\right\}$}
\RETURN First row of $\enc{V}$
\end{algorithmic}
\end{algorithm}

\section{Conclusion}
We have outlined a protocol to provide secure text analytics, by combining standard features with numeric computation and secure linear algebra with obfuscated addition. Our result may also be used for numeric, un-trusted coordinators. Whilst not guaranteeing security, the protocol addresses common issues with sharing text data -- namely visibility of identifiable information.

\end{document}